\documentclass[square,comma]{cas-dc}

\usepackage{hyperref}
\hypersetup{
    hypertex=true,
    colorlinks=true,
    linkcolor=blue,
    anchorcolor=blue,
    citecolor=blue
}

\usepackage{float}
\usepackage{placeins}
\usepackage{graphicx}
\usepackage{multirow}
\usepackage{booktabs}
\usepackage{subcaption}
\usepackage{graphicx}
\usepackage{caption}
\usepackage{subcaption}
\usepackage{times}

\def\tsc#1{\csdef{#1}{\textsc{\lowercase{#1}}\xspace}}
\tsc{WGM}
\tsc{QE}
\tsc{EP}
\tsc{PMS}
\tsc{BEC}
\tsc{DE}

\begin{document}
\let\WriteBookmarks\relax
\def\floatpagepagefraction{1}
\def\textpagefraction{.001}
\shorttitle{Jianfeng Wu et~al.}
\shortauthors{Jianfeng Wu et~al.}

\title [mode = title]{SAAT: Synergistic Alternating Aggregation Transformer for Image Super-Resolution}

\author[1]{Jianfeng Wu}
\ead{wjf5888@stu.ouc.edu.cn}

\affiliation[1]{organization={Department of Computer Science and Technology},
                addressline={Ocean University Of China}, 
                city={Qingdao},
                postcode={266100}, 
                country={China}}

\author[1]{Nannan XU}
\cormark[1]
\ead{xnn@ouc.edu.cn}

\cortext[cor1]{Corresponding author}

\begin{abstract}
Single image super-resolution is a well-known downstream task which aims to restore low-resolution images into high-resolution images. At present, models based on Transformers have shone brightly in the field of super-resolution due to their ability to capture long-term dependencies in information. However, current methods typically compute self-attention in nonoverlapping windows to save computational costs, and the standard self-attention computation only focuses on its results, thereby neglecting the useful information across channels and the rich spatial structural information generated in the intermediate process. Channel attention and spatial attention have, respectively, brought significant improvements to various downstream visual tasks in terms of extracting feature dependency and spatial structure relationships, but the synergistic relationship between channel and spatial attention has not been fully explored yet.To address these issues, we propose a novel model. Synergistic Alternating Aggregation Transformer (SAAT), which can better utilize the potential information of features. In SAAT, we introduce the Efficient Channel \& Window Synergistic Attention Group (CWSAG) and the Spatial \& Window Synergistic Attention Group (SWSAG). On the one hand, CWSAG combines efficient channel attention with shifted window attention, enhancing non-local feature fusion, and producing more visually appealing results. On the other hand, SWSAG leverages spatial attention to capture rich structured feature information, thereby enabling SAAT to more effectively extract structural features.Extensive experimental results and ablation studies demonstrate the effectiveness of SAAT in the field of super-resolution. SAAT achieves performance comparable to that of the state-of-the-art (SOTA) under the same quantity of parameters.

\end{abstract}


\maketitle

\section{Introduction}

Single image super-resolution involves reconstructing high-resolution images from low-resolution inputs. It has become a popular research area in the fields of computer vision and image restoration and has been widely applied across various domains\cite{ref1,ref2}, such as medical image super-resolution\cite{ref3,ref4}, surveillance equipment\cite{ref5}, and satellite remote sensing\cite{ref6}. Despite extensive research and applications, super-resolution models are still striving to obtain richer feature information.In the early days of 2014, the development of deep neural networks (DNNs) marked a turning point. Due to their superior performance compared to traditional methods, convolutional neural networks (CNNs) have dominated the field of super-resolution. As a result, numerous CNN-based image super-resolution techniques have been proposed\cite{ref7,ref8,ref9,ref10}.

However, due to the local processing principle of CNNs, deep neural networks are still limited in their ability to perceive global context. In recent years, following the success of natural language processing, Transformers\cite{ref11} have garnered attention in the computer vision community. Owing to the long-range dependency modeling capability of the self-attention mechanism in Transformers and their multi-scale processing advantages, Transformer-based methods have rapidly gained popularity in downstream tasks, including super-resolution. Many works have validated the effectiveness of Transformer-based approaches, such as SwinIR\cite{ref12}, SRFormer\cite{ref13}, ESRT\cite{ref14}, and HAT\cite{ref15}.

Despite the success of Transformer-based methods in the field of super-resolution, there are still several areas that need improvement. SwinIR\cite{ref12} introduced window attention, which restricts attention computation within fixed windows. This approach clearly results in a limited receptive field and an inability to fully utilize all the pixel information of the original image. SRFormer\cite{ref13} proposed a new image super-resolution permutation self-attention method, enlarging the window to obtain more information. However, the channel compression also leads to the loss of spatial information. ESRT\cite{ref14} presented a novel efficient super-resolution network for single-image super-resolution, which extracts deep features using a lightweight CNN backbone and models long-range dependencies between similar local regions in the image using a Transformer backbone. However, such feature extraction lacks flexibility and adaptability. HAT\cite{ref15} designed a new overlapping cross-attention to enhance information interaction between adjacent local windows. HAT's method of using channel attention and shifted window self-attention to obtain long-range pixel associations depends on the increase in network depth, which also increases the difficulty of model training.

As a result, when it comes to capturing local (high-frequency) information, such as edges and textures, which are finer details crucial for the global information of SR tasks, their effectiveness diminishes. The scope of global pixel dependency also needs to be further expanded. In other words, current super-resolution reconstruction networks need to simultaneously improve their ability to capture edge features and high-frequency information.

To address the aforementioned issues and further enhance the extraction of edge features and high-frequency information in images, we propose a novel super-resolution reconstruction network, SAAT, in this paper. It incorporates CNN-based efficient channel attention, window-based attention, and spatial attention mechanisms. In SAAT, we introduce a CNN-based lightweight module, the efficient channel attention module, which we combine with window attention (ECWAG). This allows us to leverage the global information perception capability of channel attention to compensate for the shortcomings of self-attention while keeping the parameter count low. Meanwhile, we employ spatial structural attention, using multi-scale, depth-shared 1D convolutions to capture multi-scale spatial information in each feature channel. This effectively integrates global context dependencies and multi-scale spatial priors while synergizing with window attention. We alternate between ECWAG and SWAG to enhance information interaction between adjacent local windows. Additionally, we utilize ConvFFN, adding a local depthwise convolution branch between the two linear layers of the FFN block to help encode more details.

Through the aforementioned model design, SAAT is able to surpass the state-of-the-art (SOTA). In summary, our contributions are as follows:
\begin{itemize}
\item[$\bullet$]We propose the Synergistic Alternating Aggregation Transformer (SAAT) for image super-resolution. SAAT aims to explore the synergistic effects between spatial attention and channel attention, enabling the capture of global channel information and spatial structural information. Our SAAT achieves significant super-resolution performance in terms of both parameter count and visual quality.
\item[$\bullet$] We introduce the Spatial \& Window Synergistic Attention Group (SWSAG), which obtains rich spatially structured feature information and effectively integrates global contextual dependencies and multi-scale spatial priors.
\item[$\bullet$]We propose the Efficient Channel \& Window Synergistic Attention Group (CWSAG), which synergizes with window-based attention and alternates with SWSAG. This achieves a balance between parameter count and computational complexity while obtaining global feature information.
\end{itemize}

\section{Related Works}

\subsection{CNN Based Super-Resolution}

Over the past period, compared with traditional super-resolution methods, CNNs have made significant progress in the field of super-resolution based on their powerful ability to extract spatial features. As a pioneer, SRCNN\cite{ref16} used a three-layer convolutional network to fit a nonlinear mapping, reconstructing low-resolution images into high-resolution ones. FSRCNN\cite{ref17} proposed a compact hourglass-shaped CNN structure to achieve faster and better image super-resolution reconstruction. ESPCN\cite{ref18} introduced sub-pixel convolution, presenting an efficient method to directly extract features on the size of low-resolution images and compute high-resolution images. However, when increasing the number of CNN layers to extract more complex features, a sharp decline in performance was observed. EDSR\cite{ref19} and RDN\cite{ref20} are SR structures based on residual networks, which addressed this issue through residual connections. SRGAN proposed using generative adversarial networks to optimize the process of generating super-resolution images. The generator of SRGAN\cite{ref21} learns the mapping from low-resolution images to high-resolution images and improves the quality of generated images through adversarial training. There are also many other works\cite{ref22,ref23} that generate realistic textures in the reconstruction process through GAN\cite{ref24}. Later, influenced by the Transformer, Wang et al.\cite{ref25} first integrated non-local attention blocks into CNNs, verifying the role of attention mechanisms in visual tasks. SAN\cite{ref26} is built on a second-order attention network, which has raised SR performance to a new level. Despite the great success of these works in the SISR field, they are also subject to significant limitations. In particular, CNNs can only extract local feature information and require a large amount of computational cost.

\subsection{Vision Transformer Based Super-Resolution}

Recently, due to the great success of Transformer\cite{ref11} in the field of natural language processing, it has also attracted the attention of researchers in the field of computer vision. With its powerful ability to capture long-range dependencies in features, Transformer has been successfully applied to various areas of computer vision, such as image classification\cite{ref27,ref28}, semantic segmentation\cite{ref29,ref30}, image restoration\cite{ref31,ref32} and so on. ViT\cite{ref33} was the first work to use Transformer to replace standard convolutions in visual tasks. Since then, a variety of Transformer-based techniques have emerged. IPT\cite{ref34} developed a video-style network and introduced multi-task pretraining for image processing. SwinIR\cite{ref12} proposed an image restoration Transformer based on a certain method. VRT\cite{ref36} introduced Transformer-based networks into video restoration. EDT\cite{ref37} adopted self-attention mechanisms and multi-related task pretraining strategies to further refresh the performance of SR. HAT\cite{ref15} added channel attention and OCAB blocks on the basis of SwinIR, and achieved SOTA with the proposed same-task pretraining strategy. These methods all applied self-attention in local regions to reduce computational complexity. However, the local design limits the use of global information, which is crucial for image SR, and the ability to capture high-frequency and low-frequency information still needs to be further expanded. SAAT is also based on Transformer networks. We propose Spatial \& Window Synergistic Attention (SWSA) and Efficient Channel \& Window Synergistic Attention (CWSA). When coordinated with window-based attention, our method integrates local and global information well and achieves good results in the task of image super-resolution.

\begin{figure*}
    \centering
    \includegraphics[width=1.0\textwidth]{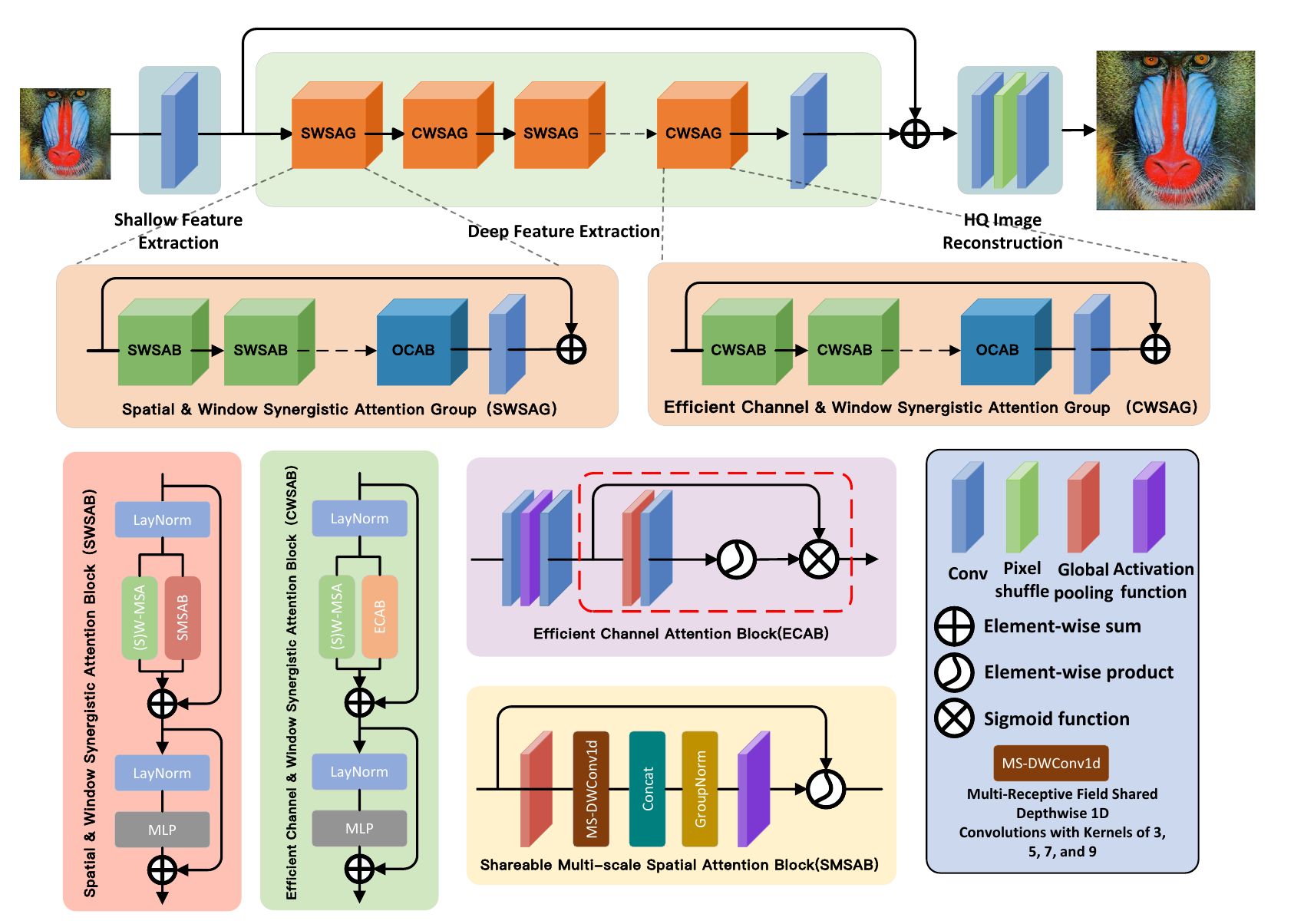}
    \caption{The overall architecture and all internal units of SAAT.} 
    \captionsetup{font=rm}
    \label{fig1}
\end{figure*}

\section{Method}
\subsection{Overall Architecture}

As shown in Figure \ref{fig1}, the overall network can be divided into three parts: shallow feature extraction, deep feature extraction, and high-resolution image reconstruction. The structural design mainly refers to mainstream works\cite{ref38,ref39}. The following is a description of the three parts:

\subsection{Shallow Feature Extraction}

Given a low-resolution image $I_{LR} \in \mathbb{R}^{H \times W \times C_{in}}$, we use a convolutional layer to extract shallow features $F_0 \in \mathbb{R}^{H \times W \times C}$, where $H$, $W$, $C_{in}$, and $C$ represent height, width, the number of input image channels, and the number of output image channels, respectively. The convolutional layer is a $3 \times 3$ convolution $H_{Conv}(\cdot)$.

\begin{equation}
F_{} = f_{sf}(I_{LR})
\end{equation}

$f_{sf}$ represents the shallow feature extraction module, which is mainly used to extract the low-frequency information of the image, providing more stable optimization and better results, while also achieving high-dimensional embedding of each pixel coordinate in the image.

\subsection{Deep Feature Extraction}

The deep feature extraction module consists of a series of Spatial \& Window Synergistic Attention Group (SWSAG) and Efficient Channel \& Window Synergistic Attention Group (CWSAG) alternating cycles, as well as a $3 \times 3$ convolutional layer $H_{Conv}(\cdot)$. $F_{sh} \in \mathbb{R}^{H \times W \times C}$ is obtained through the deep feature extraction module to $F_{dp} \in \mathbb{R}^{H \times W \times C}$, extracting the high-frequency information of the image.

\subsubsection{Spatial \& Window Synergistic Attention Group(SWSAG)}

Channel attention and spatial attention have significantly improved feature dependency and spatial structure relationship extraction for various downstream vision tasks. Although their combined use is more advantageous, the collaborative relationship between channel and spatial attention has not been fully explored. Therefore, we designed a Spatial \& Window Synergistic Attention Block (SWSAB), which includes spatial and window collaborative attention units, and incorporates an Overlapping Cross-Attention Block (OCAB) to obtain richer feature information. Finally, we added a convolutional layer with a $3 \times 3$ kernel $H_{Conv}(\cdot)$. The mathematical formula for SWSAG can be expressed as:

\begin{equation}
F_d = f_{Conv}(f_{oca}(f_{swsa}^2 \cdots (f_{swsa}^1(f_s)))) + F_s
\end{equation}

Where $f_{swsa}$, $f_{oca}$, and $f_{Conv}$ represent the spatial and window collaborative attention units, overlapping cross-attention units, and convolutional operations, respectively.

\begin{figure*}
    \centering
    \includegraphics[width=1.0\textwidth]{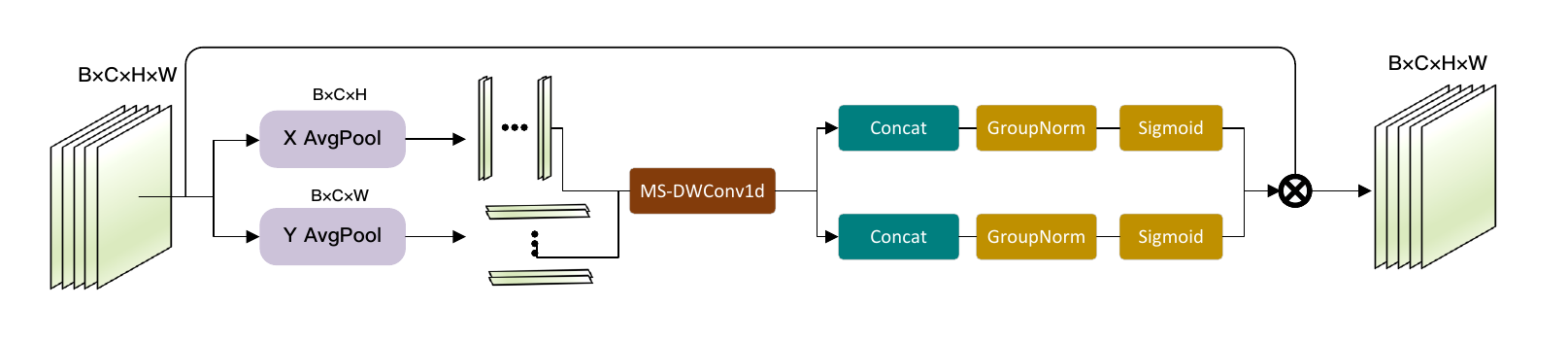}
    \caption{Shareable Multi-scale Spatial Attention Block(SMSAB).} 
    \captionsetup{font=rm}
    \label{fig2}
\end{figure*}

\textbf{Spatial \& Window Synergistic Attention Block (SWSAB):} By capturing multi-scale spatial information through shared spatial attention, it can effectively integrate global context dependencies and multi-scale spatial priors. As shown in Figure \ref{fig2}, after the first LayerNorm (LN) layer, a multi-head self-attention (S)W-MSA module and a Shared Multi-scale Spatial Attention Block (SMSAB) are inserted in parallel. The entire module's calculation can be expressed as:

\begin{equation}
F_{int} = f_{MSA}(f_{LN}(F_{in})) + \alpha f_{SMSA}(f_{LN}(F_{in})) + F_{in}
\end{equation}

\begin{equation}
F_{out} = f_{MLP}(f_{LN}(F_{in})) + F_{int}
\end{equation}

Where $F_{in}$, $F_{int}$, and $F_{out}$ are the input features, intermediate features, and output features, respectively. $f_{MSA}$, $f_{LN}$, $f_{LN}$, and $f_{SMSA}$ are (S)W-MSA, the first LayerNorm, the second LayerNorm, and SMSAB, respectively. The $\alpha$ in the formula is to avoid possible conflicts in optimization and visual representation between (S)W-MSA and SMSA.

\textbf{Shareable Multi-scale Spatial Attention Block (SMSAB):} Standard self-attention calculations only focus on the results of the computation, ignoring the rich spatial structural information generated during the intermediate process. Inspired by many insights into spatial attention\cite{ref25,ref27}, as shown in Figure 2, we first average the input $X \in \mathbb{R}^{B \times C \times H \times W}$ across each dimension to obtain two one-dimensional sequences: $X_H \in \mathbb{R}^{B \times C \times H \times W}$ and $X_W \in \mathbb{R}^{B \times C \times H \times W}$. To better learn global context dependencies and multi-scale spatial priors, $X_H$ and $X_W$ are divided into $i$ parts. Then, $K$ depth-shared 1D convolutions are applied to each part of the features to process them, extracting information at different scales. In this paper, $K$ is set to 4, and the calculation formulas are as follows:

\begin{equation}
X_H^i = DWConv1d_{\epsilon_{K_i}}^{\sigma_{K_i}}(X_H^i)
\end{equation}

\begin{equation}
X_W^i = DWConv1d_{\epsilon_{K_i}}^{\sigma_{K_i}}(X_W^i)
\end{equation}

Where the second $X_H^i$ represents the $i$-th sub-feature, with $i \in [1, K]$. Each sub-feature is independent, facilitating effective extraction of multi-scale spatial information. The first $X_H^i$ represents the spatial structural information of the $i$-th sub-feature after convolutional operations.

The features extracted are then concatenated and normalized using a GroupNorm layer. Finally, a scaling function is used to obtain spatial attention weights. The calculation process is as follows:

\begin{equation}
Attn_H = \sigma(GN_H^K(Concat(X_H^1, X_H^2, \cdots, X_H^K)))
\end{equation}

\begin{equation}
Attn_W = \sigma(GN_W^K(Concat(X_W^1, X_W^2, \cdots, X_W^K)))
\end{equation}

\begin{equation}
SMSA(X) = X_s = Attn_H \times Attn_W \times X
\end{equation}

Where $\sigma(\cdot)$ represents the Sigmoid activation function, $GN_H^K(\cdot)$ and $GN_W^K(\cdot)$ represent $K$ groups of normalization layers along $H$ and $W$ dimensions, respectively.


\textbf{Overlapping Cross-Attention Block (OCAB):} This block overlaps features between adjacent windows and establishes cross-attention between them to further enhance network performance\cite{ref15}. OCAB consists of an overlapping cross-attention layer and a Swin Transformer-like\cite{ref12} MLP layer. OCAB is implemented using a sliding window approach, dividing the input feature's dimensions $H \times W \times C$ into non-overlapping windows of size $G_0 \times G_0$ of $\frac{HW}{G^2}$. Here, $G_0$ is calculated as:

\begin{equation}
G_0 = (1 + \mu) \times G
\end{equation}

Where $\mu$ is a constant that adjusts the overlap ratio, and $G$ represents the size of the current window. During the MSA phase, queries are calculated by a linear layer, while keys and values are generated by another linear layer, followed by expansion operations. As shown in Figure 1, OCAB modules are placed at the end of the SWSAB and CWSAB modules.

\subsubsection{Efficient Channel \& Window Synergistic Attention Group(CWSAG)}

Channel attention (CA) allows the network to focus on more useful channels and adaptively reweights the features of each channel through the interdependencies between feature channels to improve learning\cite{ref41}. HAT\cite{ref15} has shown that when using channel attention, more image regions are activated due to the involvement of global information. Although CA can extract better feature information, it inevitably increases computational load, making the balance between performance and computational complexity a problem. Inspired by \cite{ref42}, we propose an efficient CA that maintains high performance while reducing computational load. The CWSAG module includes channel and window collaborative attention units, similar to SWSAG, and also incorporates OCAB and $H_{Conv}(\cdot)$.

\textbf{Efficient Channel \& Window Synergistic Attention Block (CWSAB):} Despite the significant improvements brought by spatial and channel attention in capturing feature dependencies and spatial structures, their collaborative effect has not been fully explored. Therefore, we adopt a structure that combines SWSAG and CWSAG to enhance their interaction. As shown in Figure 1, after the first LayerNorm (LN) layer, a multi-head self-attention (S)W-MSA module is inserted in parallel with an Efficient Channel Attention Block (ECA). The entire module's computation can be expressed as:

\begin{equation}
F_{int} = f_{MSA}(f_{LN}^1(F_{in})) + \beta f_{ECA}(f_{LN}^2(F_{in})) + F_{in}
\end{equation}

\begin{equation}
F_{out} = f_{MLP}(f_{LN}^2(F_{int})) + F_{int}
\end{equation}

Where $F_{in}$, $F_{int}$, and $F_{out}$ are the input features, intermediate features, and output features, respectively. $f_{MSA}$, $f_{LN}^1$, $f_{LN}^2$, and $f_{ECA}$ are (S)W-MSA, the first LayerNorm, the second LayerNorm, and ECAB, respectively. The $\beta$ in the formula is to avoid possible conflicts between (S)W-MSA and ECAB in optimization and visual representation.

\textbf{Efficient Channel Attention Block (ECA):} The structure of ECAB is shown in Figure 1. ECAB mainly consists of two standard convolutional layers, one with an activation function and one efficient channel attention (ECA) module. The input to ECAB can be expressed as:

\begin{equation}
F_{int} = f_{Conv2}(f_{act}(f_{Conv1}(F_{in})))
\end{equation}

\begin{equation}
F_{out} = f_{Sig}(f_{Conv3}^k(f_{Avg}(F_{int}))) + F_{int}
\end{equation}

Where $f_{Conv1}(\cdot)$, $f_{Conv2}(\cdot)$, and $F_{int}$ represent the first convolutional layer, the second convolutional layer, and the intermediate features of the compressed features, respectively. $f_{Avg}(\cdot)$, $f_{Conv3}^k(\cdot)$, $f_{Sig}(\cdot)$, and $F_{out}$ represent the average pooling layer, the third convolutional layer with kernel size $k$, the Sigmoid activation function, and the output of the compressed features, respectively. ECAB also employs a local cross-channel interaction strategy without dimensionality reduction, implemented by a fast one-dimensional convolution with kernel size $k$, where $k$ represents the range of local cross-channel interactions, i.e., how many fields participate in the attention prediction of a channel. Given a compressed feature $C$, $k$ can be adaptively expressed as:

\begin{equation}
k = \Psi(C) = \left\lfloor \frac{\log_2(C) + b}{\gamma} \right\rfloor_{odd}
\end{equation}

Where $\Psi(\cdot)$ is the mapping function, $\gamma$ and $b$ are the parameters of the mapping function. In this paper, we set $\gamma$ and $b$ to 2 and 1, respectively, in all experiments. $\lfloor t \rfloor_{odd}$ represents the nearest odd integer to $t$. Through the mapping function, high-dimensional channels have a longer range of interaction, while low-dimensional channels have a shorter distance through nonlinear mapping.

\section{Experiments}

\subsection{Datasets}

We employ two extensive datasets, DIV2K\cite{ref43} and Flicker2K\cite{ref44}, to train the model in order to prevent overfitting that might occur if trained solely on DIV2K. The model is then evaluated on standard super-resolution benchmarks: Set5\cite{ref45}, Set14\cite{ref46}, BSD100\cite{ref47}, Urban100\cite{ref48}, and Manga109\cite{ref49}.

\begin{table*}[t]
\centering
\caption{Performance comparison of super-resolution methods across different scales and datasets. PSNR (dB) and SSIM metrics are reported.}
\label{tab:comparison}
\resizebox{\textwidth}{!}{%
\begin{tabular}{@{}lcccccccccccc@{}}
\toprule
\multirow{2}{*}{Method} & \multirow{2}{*}{Scale} & \multirow{2}{*}{Training Dataset} & \multicolumn{2}{c}{Set5} & \multicolumn{2}{c}{Set14} & \multicolumn{2}{c}{BSD100} & \multicolumn{2}{c}{Urban100} & \multicolumn{2}{c}{Manga109} \\
 & & & PSNR & SSIM & PSNR & SSIM & PSNR & SSIM & PSNR & SSIM & PSNR & SSIM \\
\midrule
EDSR & ×2 & DIV2K & 38.11 & 0.9602 & 33.92 & 0.9195 & 32.32 & 0.9013 & 32.93 & 0.9351 & 39.10 & 0.9773 \\
RCAN & ×2 & DIV2K & 38.27 & 0.9614 & 34.12 & 0.9216 & 32.41 & 0.9027 & 32.34 & 0.9384 & 39.44 & 0.9786 \\
SAN & ×2 & DIV2K & 38.31 & 0.9620 & 34.07 & 0.9213 & 32.42 & 0.9028 & 33.10 & 0.9370 & 39.32 & 0.9792 \\
IGNN & ×2 & DIV2K & 38.24 & 0.9613 & 34.07 & 0.9217 & 32.41 & 0.9025 & 33.23 & 0.9383 & 39.35 & 0.9786 \\
HAN & ×2 & DIV2K & 38.27 & 0.9614 & 34.16 & 0.9217 & 32.41 & 0.9027 & 33.35 & 0.9385 & 39.46 & 0.9785 \\
NLSN & ×2 & DIV2K & 38.34 & 0.9618 & 34.08 & 0.9231 & 32.43 & 0.9027 & 33.42 & 0.9394 & 39.59 & 0.9789 \\
SwinIR & ×2 & DIV2K+Flickr2K & 38.42 & 0.9623 & 34.46 & 0.9250 & 32.53 & 0.9041 & 33.81 & 0.9427 & 39.92 & 0.9797 \\
SRFormer & ×2 & DIV2K+Flickr2K & 38.51 & 0.9627 & 34.44 & 0.9253 & 32.57 & 0.9046 & 34.09 & 0.9449 & 40.07 & 0.9802 \\
EDT & ×2 & DIV2K+Flickr2K & 38.45 & 0.9624 & 34.57 & 0.9258 & 32.52 & 0.9041 & 33.80 & 0.9425 & 39.93 & 0.9800 \\
HAT & ×2 & DIV2K+Flickr2K & 38.63 & 0.9630 & 34.86 & \textcolor{red}{0.9274} & 32.62 & 0.9053 & 34.45 & \textcolor{red}{0.9466} & 40.26 & 0.9809 \\
SAAT(ours) & ×2 & DIV2K+Flickr2K & \textcolor{red}{38.71} & \textcolor{red}{0.9637} & \textcolor{red}{34.94} & 0.9273 & \textcolor{red}{32.65} & \textcolor{red}{0.9057 }& \textcolor{red}{34.50} & 0.9462 & \textcolor{red}{40.44} & \textcolor{red}{0.9814} \\

\midrule
EDSR & ×3 & DIV2K & 34.65 & 0.9280 & 30.52 & 0.8462 & 29.25 & 0.8093 & 28.80 & 0.8653 & 34.17 & 0.9476 \\
RCAN & ×3 & DIV2K & 34.74 & 0.9299 & 30.65 & 0.8482 & 29.32 & 0.8111 & 29.09 & 0.8702 & 34.44 & 0.9499 \\
SAN & ×3 & DIV2K & 34.75 & 0.9300 & 30.59 & 0.8476 & 29.33 & 0.8112 & 28.93 & 0.8671 & 34.30 & 0.9494 \\
IGNN & ×3 & DIV2K& 34.72 & 0.9298 & 30.66 & 0.8484 & 29.31 & 0.8105 & 29.03 & 0.8696 & 34.39 & 0.9496 \\
HAN & ×3 & DIV2K & 34.75 & 0.9299 & 30.67 & 0.8483 & 29.32 & 0.8110 & 29.10 & 0.8705 & 34.48 & 0.9500 \\
NLSN & ×3 & DIV2K & 34.85 & 0.9306 & 30.70 & 0.8485 & 29.34 & 0.8117 & 29.25 & 0.8726 & 34.57 & 0.9508 \\
SwinIR & ×3 & DIV2K+Flickr2K & 34.97 & 0.9318 & 30.93 & 0.8534 & 29.46 & 0.8145 & 29.75 & 0.8826 & 35.12 & 0.9537 \\
SRFormer & ×3 & DIV2K+Flickr2K & 34.42 & 0.9268 & 30.43 & 0.8433 & 29.15 & 0.8063 & 28.46 & 0.8574 & 33.95 & 0.9455 \\
EDT & ×3 & DIV2K+Flickr2K & 35.02 & 0.9323 & 30.94 & 0.8540 & 29.48 & \textcolor{red}{0.8156} & 30.04 & 0.8865 & 35.26 & 0.9543 \\
HAT & ×3 & DIV2K+Flickr2K & 34.97 & 0.9316 & 30.89 & 0.8527 & 29.44 & 0.8142 & 29.72 & 0.8814 & 35.13 & 0.9534 \\
SAAT(ours) & ×3 & DIV2K+Flickr2K & \textcolor{red}{35.17} & \textcolor{red}{0.9336} & \textcolor{red}{31.15} & \textcolor{red}{0.8563} & \textcolor{red}{29.59} & 0.8166 & \textcolor{red}{30.45} & \textcolor{red}{0.8903} & \textcolor{red}{35.59} & \textcolor{red}{0.9554} \\

\midrule
EDSR & ×4 & DIV2K & 32.46 & 0.8968 & 28.80 & 0.7876 & 27.71 & 0.7420 & 26.64 & 0.8033 & 31.02 & 0.9148 \\
RCAN & ×4 & DIV2K & 32.63 & 0.9002 & 28.87 & 0.7889 & 27.77 & 0.7436 & 26.82 & 0.8087 & 31.22 & 0.9173 \\
SAN & ×4 & DIV2K & 32.64 & 0.9003 & 28.92 & 0.7888 & 27.78 & 0.7436 & 26.79 & 0.8068 & 31.18 & 0.9169 \\
IGNN & ×4 & DIV2K & 32.57 & 0.8998 & 28.85 & 0.7891 & 27.77 & 0.7434 & 26.84 & 0.8090 & 31.28 & 0.9182 \\
HAN & ×4 & DIV2K & 32.64 & 0.9002 & 28.90 & 0.7890 & 27.80 & 0.7442 & 26.85 & 0.8094 & 31.42 & 0.9177 \\
NLSN & ×4 & DIV2K & 32.59 & 0.9000 & 28.87 & 0.7891 & 27.78 & 0.7444 & 26.96 & 0.8109 & 31.27 & 0.9184 \\
SwinIR & ×4 & DIV2K+Flickr2K & 32.92 & 0.9044 & 29.09 & 0.7950 & 27.92 & 0.7489 & 27.45 & 0.8254 & 32.03 & 0.9260 \\
SRFormer & ×4 & DIV2K+Flickr2K & 32.19 & 0.8947 & 28.69 & 0.7833 & 27.69 & 0.7379 & 26.39 & 0.7962 & 30.75 & 0.9100 \\
EDT & ×4 & DIV2K+Flickr2K & 32.93 & 0.9041 & 29.08 & 0.7953 & 27.94 & 0.7502 & 27.68 & 0.8311 & 32.21 & 0.9271 \\
HAT & ×4 & DIV2K+Flickr2K & 32.82 & 0.9031 & 29.09 & 0.7939 & 27.91 & 0.7483 & 27.46 & 0.8246 & 32.05 & 0.9254 \\
SAAT(ours) & ×4 & DIV2K+Flickr2K & \textcolor{red}{33.11} & \textcolor{red}{0.9060} & \textcolor{red}{29.32} & \textcolor{red}{0.7982} & \textcolor{red}{28.05} & \textcolor{red}{0.7520} & \textcolor{red}{28.12} & \textcolor{red}{0.8391} & \textcolor{red}{32.57} & \textcolor{red}{0.9298} \\

\bottomrule
\end{tabular}%
}
\end{table*}

\subsection{Experimental Settings}

We employ geometric methods such as flipping or rotating for data augmentation and then crop the augmented images into 64×64 patches. For the structure of SAAT, we alternately arrange three Spatial \& Window Synergistic Attention Groups (SWSAGs) and three Efficient Channel \& Window Synergistic Attention Groups (CWSAGs). Within each SWSAG and CWSAG, we set four Spatial \& Window Synergistic Attention Blocks (SWSABs) and four Efficient Channel \& Window Synergistic Attention Blocks (CWSABs), respectively. Each SWSAG and CWSAG unit represents a unique shift size, namely 0, 8, 16, and 24. The number of channels, attention heads, and window size are set to 180, 6, and 16, respectively. For other hyperparameters, we will discuss them in the ablation study. The total number of training iterations and batch size are set to 500K and 32, respectively. The learning rate is initialized to 2e-4 and is halved at [250K, 400K, 450K, 475K]. We use L1 loss for model training. We adopt a pre-trained ×2 model and fine-tune it to ×4 resolution. We use the ADAM\cite{ref50} optimizer to train the model with parameters $\beta_1$ and $\beta_2$.

\subsection{Ablation Study}
\subsubsection{Effectiveness of SMSAB and ECAB}

We have demonstrated the effectiveness of the SMSAB and ECAB modules proposed in this paper through experiments. Experiments were conducted on the Urban100\cite{ref48} dataset to evaluate PSNR/SSIM. The results are shown in Table \ref{tab:ablation}. The best performance was achieved when using both modules simultaneously. In comparison, the performance gains obtained by using either the SMSAB or ECAB module alone were not as good as when they were used together. Although the performance of the SMSAB module alone was slightly better than that of the ECAB module alone, the ECAB module also reduced computational load while maintaining high performance. This means that our proposed method not only performs well in terms of PSNR and visual quality of the restored images but also has favorable computational costs.

\begin{table}[pos=htbp]
\centering
\caption{Ablation study on SMSAB and ECAB}
\label{tab:ablation}
\scriptsize 
\setlength{\tabcolsep}{3.5pt} 
\begin{tabular}{@{}lcccc@{}}
\toprule
 & \multicolumn{4}{c}{Baseline} \\
\cmidrule(lr){2-5}
SMSAB & $\times$ & $\checkmark$ & $\times$ & $\checkmark$ \\
ECAB & $\times$ & $\times$ & $\checkmark$ & $\checkmark$ \\
\midrule
PSNR/SSIM & 27.23/0.8242 & 28.35/0.8376 & 28.16/0.8371 & 28.47/0.8459 \\
\bottomrule
\end{tabular}
\end{table}

\subsubsection{Effects of different designs of SWSAB and ECAB}
To examine the impact of the weight factors \( \alpha \) and \( \beta \), we measured the model outcomes for four different values of \( \alpha \) and \( \beta \) (0, 1, 0.1, 0.01). As shown in the table \ref{tab:alpha_beta}, these experiments determined the optimal values for \( \alpha \) and \( \beta \) to be 0.01 and 0.01, respectively.

\begin{table}[pos=htbp]
\centering
\caption{Ablation study on weight factors $\alpha$ and $\beta$}
\label{tab:alpha_beta}
\begin{tabular}{ccccc}
\toprule
$\alpha$ & 0 & 1 & 0.1 & 0.01 \\
\midrule
PSNR (dB) & 27.73 & 27.68 & 27.79 & 27.85 \\
\bottomrule
\end{tabular}
\end{table}

\vspace{-10pt} 

\begin{table}[pos=htbp]
\centering
\label{tab:beta}
\begin{tabular}{ccccc}
\toprule
$\beta$ & 0 & 1 & 0.1 & 0.01 \\
\midrule
PSNR (dB) & 27.78 & 27.73 & 27.81 & 27.83 \\
\bottomrule
\end{tabular}
\end{table}

\subsubsection{Overlapping Constant}

In the OCAB, we use a constant $\mu$ to determine the overlap range between two consecutive windows during cross-attention. We adjust the $\mu$ value from 0 to 0.75 to study the impact of various overlap ratios, and the results are shown in the table. Here, $\mu = 0$ corresponds to a standard Transformer block. The results shown in the table \ref{tab:mu_ablation} indicate that the model provides optimal performance at $\mu = 0.5$.

\begin{table}[pos=htbp]
\centering
\caption{Ablation study on OCAB overlap ratio $\mu$}
\captionsetup{font=rm}
\label{tab:mu_ablation}
\begin{tabular}{ccccc}
\toprule
$\mu$ & 0 & 0.25 & 0.5 & 0.75 \\
\midrule
PSNR (dB) & 27.75 & 27.78 & 27.85 & 27.78 \\
\bottomrule
\end{tabular}
\end{table}

\begin{figure*}
    \centering
    \includegraphics[width=1.0\textwidth]{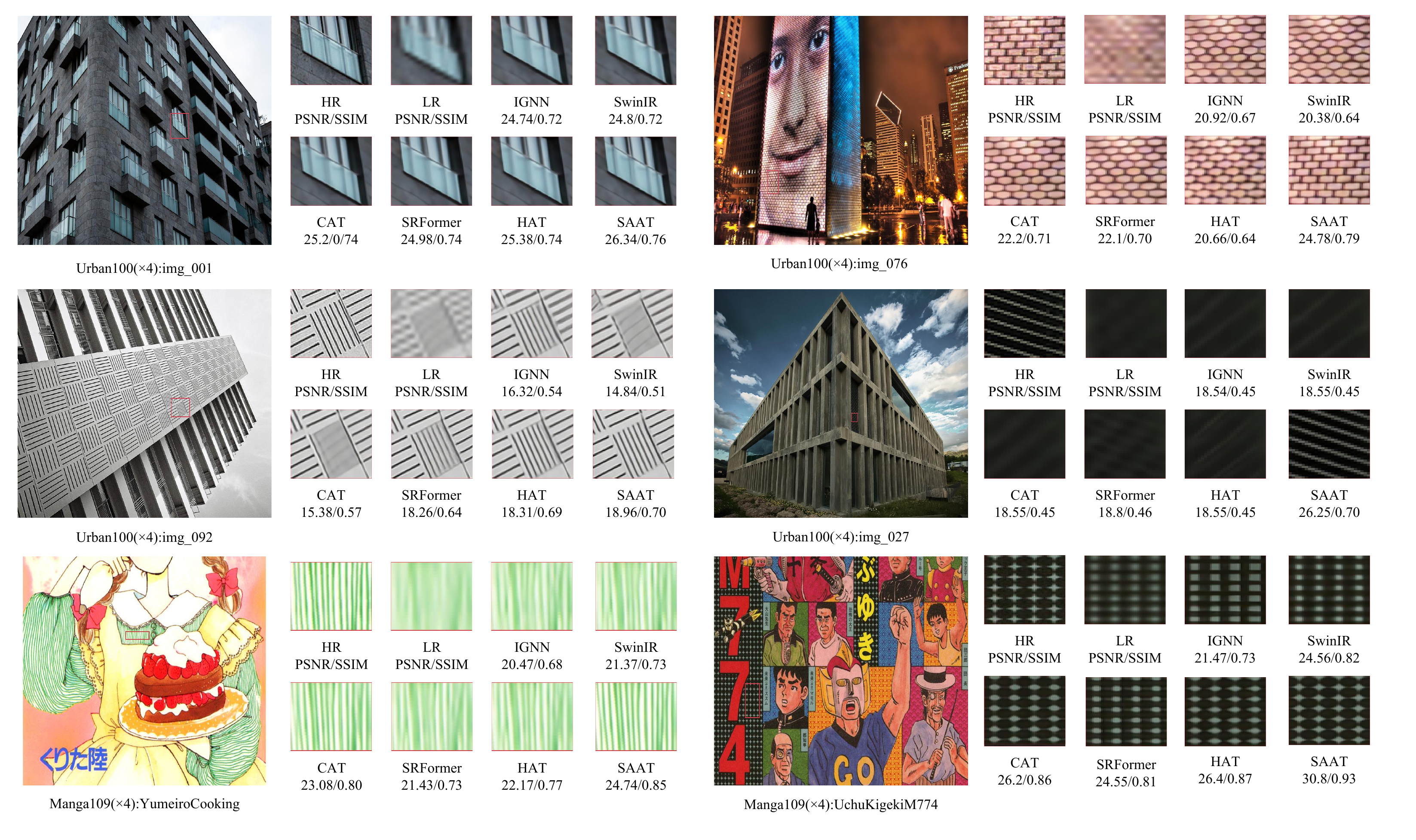}
    \caption{Visual comparison on ×4 SR. PSNR/SSIM is calculated in patches marked with red boxes in the images.} 
    \captionsetup{font=rm}
    \label{fig3}
\end{figure*}

\subsection{Comparison with State-of-the-Art Methods}
\subsubsection{Quantitative comparison}

The table shows the comparison of our model with the following models in terms of PSNR and SSIM: EDSR\cite{ref19}, RCAN\cite{ref51}, SAN\cite{ref52}, IGNN\cite{ref53}, HAN\cite{ref54}, NLSN\cite{ref55}, SwinIR\cite{ref12}, SRFormer\cite{ref13}, EDT\cite{ref56}, and HAT\cite{ref15}. It can be seen that our method achieves the best performance on almost all scales across five datasets. Specifically, SAAT outperforms SwinIR on all scales. Particularly on Urban100\cite{ref48} and Manga109\cite{ref49}, which contain a large number of repetitive textures, SAAT improves by 0.10dB compared to SwinIR. It should be noted that HAT introduces channel attention into the model. However, the performance of HAT\cite{ref15} is not as good as SAAT, which proves the effectiveness of our proposed method.

\subsubsection{Visual comparison}

We provide some visual comparison results in the Figure \ref{fig3}. The comparison results are selected from the Urban100 and Manga109 datasets: "img001", "img027", "img076","img092", "UchuKigekiM774", and "YumeiroCooking". In the figure, PSNR and SSIM are calculated from the patches marked with red boxes in the images. Visually, SAAT can better restore the texture details of the images. Compared to other advanced methods, SAAT recovers the edges of the images more clearly. When recovering images "img027" and "img076" with other state-of-the-art methods, we can see many blurry areas, while SAAT produces good visual effects. The comparison of visual effects indicates that our proposed method also achieves superior performance.

\section{Conclusion}
This work introduces a novel Synergistic Alternating Aggregation Transformer (SAAT) for single-image super-resolution. The model combines convolutional neural network-based attention mechanisms with the Swin Transformer, effectively leveraging high-frequency feature information in images to more accurately restore details and edges. By integrating channel attention, spatial attention, and window attention, the network enhances multi-level structural similarity. Trained on the DF2K dataset and validated on the Set5, Set14, BSD100, Urban100, and Manga109 datasets, the model demonstrates superior performance, outperforming state-of-the-art techniques on benchmark datasets for single-image super-resolution tasks.

\end{document}